\documentclass{llncs}

\usepackage[utf8]{inputenc}
\usepackage{amssymb}
\usepackage{amsfonts}

\usepackage{multirow,multicol}
\usepackage{color,tabularx, colortbl,amsmath,bm,arydshln}
\usepackage{enumitem,kantlipsum,arydshln}
\usepackage{booktabs,url,tcolorbox,mathtools} 
\usepackage{newtxtext,newtxmath}
\usepackage[]{graphicx}
\usepackage{subfigure}
\usepackage{caption, subcaption}
\usepackage{changepage}
\usepackage{graphicx}
\usepackage{filecontents}
\usepackage{multirow,multicol}
\usepackage{color,tabularx, colortbl,amssymb,amsmath,bm,arydshln}
\usepackage{enumitem,kantlipsum,arydshln}
\usepackage{booktabs,url,tcolorbox,mathtools} 
\usepackage{xcolor}
\usepackage[algo2e,linesnumbered,ruled,vlined]{algorithm2e}
\usepackage{booktabs}
\usepackage{subcaption}
\usepackage{hyperref}
\usepackage[T1]{fontenc}
\usepackage{amsmath}
\newcommand{\ds}{\texttt{PromptSET}\xspace}
\newcommand{\orcid}[1]{\href{https://orcid.org/#1}{\includegraphics[width=8pt]{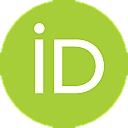}}}

\begin{document}
\title{ Benchmarking Prompt Sensitivity in\\ Large Language Models}
\titlerunning{Benchmarking Prompt Sensitivity in Large Language Models}

\author{Amirhossein Razavi$^*$ \inst{1} \orcid{0009-0007-7052-5100} \and
Mina Soltangheis$^*$ \inst{1,2} \orcid{0009-0006-1246-7363} \and
Negar Arabzadeh\inst{3} \orcid{0000-0002-4411-7089} \and \\
Sara Salamat\inst{1} \orcid{0009-0007-3676-6023}\and 
Morteza Zihayat\inst{1} \orcid{0000-0002-1144-7364} \and
Ebrahim Bagheri\inst{4} \orcid{0000-0002-5148-6237}}

\authorrunning{F. Author et al.}

\institute{Toronto Metropolitan University, Toronto ON, Canada
\email{\{amirhossein.razavi,msoltangheis,sara.salamat,mzihayat\}@torontomu.ca}\and
Wondeur Ai, Toronto ON, Canada \\
\email{mina@wondeur.ai}\and
University of Waterloo, Waterloo ON, Canada \\
\email{narabzad@uwaterloo.ca}\and
University of Toronto, Toronto ON, Canada \\
\email{ebrahim.bagheri@utoronto.ca}
}

%
%
%
%
%
%
\maketitle              
\def\thefootnote{*}\footnotetext{These authors contributed equally to this work and are listed in alphabetical order.}
\begin{abstract}
Large language Models (LLMs) are highly sensitive to variations in prompt formulation, which can significantly impact their ability to generate accurate responses. In this paper, we introduce a new task, Prompt Sensitivity Prediction, and a dataset \ds designed to investigate the effects of slight prompt variations on LLM performance. Using TriviaQA and HotpotQA datasets as the foundation of our work, we generate prompt variations and evaluate their effectiveness across multiple LLMs. We benchmark the \textit{prompt sensitivity prediction} task employing state-of-the-art methods from related tasks, including LLM-based self-evaluation, text classification, and query performance prediction techniques. Our findings reveal that existing methods struggle to effectively address prompt sensitivity prediction, underscoring the need to understand how information needs should be phrased for accurate LLM responses.
\end{abstract}

%
%
\section{Introduction}
Large language models (LLMs) can generate human-like responses to a wide array of prompts, from answering specific queries to planning for accumulating information to answer complex questions \cite{Kamalloo2023}. Despite their usefulness, a notable challenge in working LLMs is their sensitivity to prompt formulation \cite{Loya2023}.
Small variations in the phrasing, structure, or even punctuation of prompts can often lead to substantially different outputs~\cite{sclar2023quantifying,Raj2023Semantic,Mu2023Navigating}. To illustrate this issue, consider the sample prompts shown in Table \ref{table:prompt_sensitivity_examples}. In this table, we present samples from the TriviaQA \cite{joshi2017triviaqa} and HotPotQA \cite{yang2018hotpotqa} question-answering datasets where the LLM responds correctly and accurately to the original prompts. However, with only slight modifications in wording, we observe that the LLM (in this case LLaMA3.1) fails to provides the correct response. 

This challenge, which we refer to as \textit{prompt sensitivity}, highlights the challenges users face when crafting their prompts~\cite{Feng2024UnveilingAM,zhuo2024prosa}. For this reason, \textit{prompt engineering}, the art of designing effective prompts, has become an active area of research~\cite{Lo2023The,Bhargava2023magicword}. 
Researchers have already examined the effects of various prompt modifications, including minor structural and formatting changes \cite{sclar2023quantifying}, adversarial prompting \cite{Zhu2023PromptBench}, and generating prompts with different levels of specificity \cite{Murr2023Testing}. We hypothesize that prompt  sensitivity could arise because of several reasons. For instance, an LLM may successfully respond to prompts closely aligned with examples seen during training, but struggle with slight modifications that it has not encountered in the past \cite{arabzadeh-etal-2024-assessing}. Another factor could be the model’s reliance on specific syntactic or semantic patterns to interpret prompts accurately, which may be impacted due to slight changes in the prompt.


To this end and in this paper, we introduce a novel task and its accompanying dataset specifically curated for \textit{prompt sensitivity prediction}.  By curating a collection of prompts and their variations, we aim to predict whether a given LLM would be able to effectively respond to an input prompt or whether it would fail to provide a satisfactory response. Our proposed dataset serves as a benchmark for studying prompt sensitivity, thus setting the stage for forthcoming studies in prompt engineering and the evaluation of LLM responses to prompt variations.

\begin{table}[t]
\centering
\caption{Samples of sensitive prompts from HotpotQA and TriviaQA datasets.}
\scalebox{0.68}{
\begin{tabular}{p{1.4cm}p{5.7cm}p{0.1cm}p{5.39cm}p{0.1cm}p{1.4cm}p{1.6cm}p{1.2cm}}
\hline \hline
\textbf{Dataset} & \textbf{Original Prompt} & & \textbf{Alternative Prompt} & & \textbf{Original Answer} & \textbf{Alternative Answer} & \textbf{Correct Answer} \\
\hline
\small HotpotQA & What American actor and comedian known for playing the role of Newman in Seinfeld, also stars in the series The Exes on TV Land? & &What is the name of the American actor who played Newman in Seinfeld and appears in TV Land's comedy series The Exes & & Wayne Knight & Jerry Seinfeld co-star & Wayne Knight \\[2em]
TriviaQA & At which city do the Blue and White Niles meet? & & At which geographical location do the Blue and White Niles meet & & Sudan's confluence & Khartoum & Khartoum \\
\hline \hline
\end{tabular}}
\label{table:prompt_sensitivity_examples}
\end{table}


A common prompt engineering strategy is to ask the LLM itself to reformulate the prompt in a way that a more desirable output would be generated for the revised prompt by the LLM \cite{hosseini2024enhanced,bigdeli2024learning}.  While LLMs can autonomously generate different prompt variations, they cannot assess which variations are most effective, pointing to the fact that LLMs themselves are oblivious to the representation of optimal prompt variations. 
To establish a benchmark for this challenge, we formally introduce the \textit{Prompt Sensitivity Prediction} task, which is concerned with assessing the effectiveness of a user prompt and its variations. We systematically curate our dataset based on the TriviaQA and HotpotQA \cite{joshi2017triviaqa,yang2018hotpotqa}, which consist of prompts that have deterministic and  concise answers. 
To benchmark this task, we draw parallels with established tasks in text classification (TC) ~\cite{gasparetto2022survey,CollinsThompson2010} and query performance prediction (QPP) ~\cite{meng2024query,Arabzadeh2020,Hambarde2023,arabzadeh2024query,arabzadeh2024queryv2}, as they share resemblance with prompt sensitivity prediction. 
Our experiments show that such baselines fail to perform effectively for this task, underscoring the need for novel approaches tailored in particular for \textit{prompt sensitivity prediction}. In summary, the contributions of our work in this paper include: \textbf{(1)} We define the prompt sensitivity prediction task, outlining the requirements and challenges involved in identifying effective prompts; \textbf{(2)} We introduce and publicly release a comprehensive dataset for Prompt Sensitivity Evaluation Task (\ds), focusing on slight prompt modifications that unveil LLM  sensitivity to prompt variations\footnote{\url{https://github.com/Narabzad/prompt-sensitivity}}; and, \textbf{(3)} We benchmark the prompt sensitivity prediction task using state-of-the-art methods, including text classification, query performance prediction, and LLM-based baselines, to highlight the complexity of the proposed task.

\section{Methodology}

\noindent \textbf{The Task Definition.} Our proposed task of \textit{Prompt Sensitivity Prediction} aims to predict whether a given prompt can be effectively fulfilled by the LLM whose response to the prompt would satisfy the users' information need. More specifically, given a prompt $p$ with a specific information need $I_p$, we consider a set of similar prompts, denoted $\mathcal{P}=\{ p' | Sim \big< p,p'\big> > \tau \ \text{and } I_p == I_{p'}\}$, where each variation $p'$ shares the same information need $I_p$ and maintains a similarity with $p$ above a predefined threshold $\tau$. These prompts $\{ p' \}$ are designed to be only slightly modified versions of $p$, ensuring they still reflect the same information need of the user. The goal of this task is to predict, for a given prompt $p_i$, whether the LLM will generate a response that accurately respond to the underlying information need ${I_p}_i$. 

\noindent \textbf{The Dataset for the Task.} To create the gold standard dataset for the prompt sensitivity task, we adopt a systematic process to generate prompt variations and evaluate their effectiveness as follows:

\begin{enumerate}
    \item \textbf{Selecting Prompts}: We start by choosing a set of initial prompts, denoted as $\mathcal{P}$, where each prompt $p \in \mathcal{P}$ is seeking a distinct information need $I_p$.
    \item \textbf{Generating Variations}: For each prompt $p$ in the set, we use an LLM $\mathcal{L}$ to generate $N$ variations $p'  = \mathcal{L}(p \mid I_p = I_{p'})$. Here, $\mathcal{L}(p)$ denotes the process of generating variations of prompt $p$, where each variation $p'$ retains high semantic similarity with $p$, i.e., $( Sim \big< p,p'\big> > \tau )$ and preserves the original information need $I_p$.
    \item \textbf{Filtering Variations}: We process and filter out any generated variations $p'$ that do not meet specific criteria for similarity and alignment with the original prompt $p$ including LLM hallucinated content. 
    \item \textbf{LLM Response Generation}:
    For each prompt $p$ and its variations $\{ p' \in \mathcal{P'} \}$, we ask the LLM to respond to the prompt, denoted $a_p \in \mathcal{A_P}$ for the original prompt and $a_{p'} \in \mathcal{A'_P}$ for each variation.
\end{enumerate}
The combination of $\mathcal{P} \cup \mathcal{P'}$  as well as their LLM generated answers $\mathcal{A_P} \cup \mathcal{A'_P}$ form the \ds dataset for this task. 

\noindent \textbf{Source Data.} To build our dataset, we require a set of prompts that have  human annotated answers available as well as having reliable evaluation with deterministic results. Therefore, we selected two widely-used question-answering datasets, TriviaQA \cite{joshi2017triviaqa} and HotpotQA datasets \cite{yang2018hotpotqa} that meet these requirements. TriviaQA is a reading comprehension dataset containing over 650K question-answer-evidence triples. The questions are on average 14 words. Each question has a collection of accepted answers including a list of aliases and normalized version of the answers; the majority of which are specific and short \cite{joshi2017triviaqa}. 
Furthermore, HotpotQA is a large-scale question-answering dataset consists of 113k training question-answer pairs. Unlike TriviaQA, HotpotQA includes complex multi-hop and comparison questions that require reasoning across multiple documents to answer accurately \cite{yang2018hotpotqa}.

In our work, we randomly sampled 12K questions from the train set of each of these datasets paired with their provided answers. After removing questions that were less than 4 words or more than 40 words, we were able to obtain 11,469 unique questions and their respective answers from these datasets. We split these questions using a 70-30 ratio for training and testing, i.e., 8,028 and 3,441 questions, respectively. We consider each of these questions to be prompts that would be submitted to an LLM for a response.
	
\noindent \textbf{Generating Prompt Variations.} For each of the 11,469 unique prompts in our dataset, we generate several prompt variations. The objective of this step to generate slight variations of the original prompt, each of which may or may not be satisfiable by the LLM. 
To generate prompt variations, we utilize two widely-used LLMs that have demonstrated strong performance across many downstream tasks, namely the pre-trained LLaMA 3.1 with 8B parameters \cite{LLaMA3} and Mistral-nemo \cite{jiang2023mistral7b}. We chose open-source models to ensure reproducibility and facilitate further research in \textit{prompt sensitivity}. We designed the instructions for the LLMs to be as clear and straightforward as possible, similar to those intended for human use. The primary goal was to generate variations that retain the same semantics of the original prompt, instructing the model to produce a rephrased prompt that does not answer the question directly but maintains the same information need and semantic content
\footnote{Due to space constraints, we have provided the full set of instructions in our GitHub repository.}.
However, while instructed explicitly, LLMs occasionally deviate from the instructions due to hallucination. To address this, we filter out prompts that did not have at least nine valid variations generated and excluded prompts with fewer than four terms to maintain quality and consistency across the dataset. At the end, our dataset consists of 11,469 prompts, each with 9 different variations, resulting in 114,690K variations. We ran each of these prompts (the original prompt and its nine variations) against the LLMs and generated an answer for each. We then compared the produced answer against the expected answer in the TriviaQA and HotpotQA datasets. Each prompt or prompt variations were labeled as being answerable by the LLMs depending on the answer they generated and whether it aligned with the expected answer. On this basis, the objective of the \textit{Prompt Sensitivity Prediction} task is to predict whether an LLM can correctly answer an input prompt. 

\begin{table}[t]
\centering
\caption{Results of baselines on \ds.}
\scalebox{0.9}{\begin{tabular}{p{0.8cm}llrrrrrrrr}
\hline
\hline
\multirow{2}{*}{} & \multirow{2}{*}{Category} & \multirow{2}{*}{Method} & \multicolumn{4}{c}{\ds -TriviaQA} & \multicolumn{4}{c}{\ds - HotPotQA} \\
 &  &  & Accuracy & F1 & Recall & Precision & Accuracy & F1 & Recall & Precision \\ \hline
\multirow{10}{*}{\rotatebox{90}{\centering Mistral Answers}} & \multirow{2}{*}{LLM-Based} & Mistral & 0.5045 & 0.5858 & 0.7743 & 0.4711 & 0.3735 & 0.2005 & 0.6912 & 0.1173 \\
 &  & LLaMA & 0.4656 & 0.6239 & 0.9798 & 0.4577 & 0.1696 & 0.2050 & 0.9419 & 0.1150 \\ \cline{2-11} 
 & {Text Classification}   & BERT & 0.660 & 0.659 & 0.620 & 0.654 & 0.526 & 0.360 & 0.017 & 0.813 \\ \cline{2-11} 
 &  & CC & 0.506 & 0.453 & 0.452 & 0.454 & 0.549 & 0.209 & 0.524 & 0.130 \\
 & Specificity-based  & DC & 0.484 & 0.448 & 0.463 & 0.434 & 0.565 & 0.199 & 0.475 & 0.126 \\
 & QPP  & IEF & 0.505 & 0.462 & 0.469 & 0.455 & 0.535 & 0.204 & 0.526 & 0.127 \\
 &  & PageRank & 0.481 & 0.444 & 0.458 & 0.431 & 0.533 & 0.153 & 0.370 & 0.096 \\ \cline{2-11} 
 & Supervised QPP & BERTPE & 0.648 & 0.627 & 0.644 & 0.611 & 0.710 & 0.318 & 0.594 & 0.217 \\ \hline
 \hline
\multirow{10}{*}{\rotatebox{90}{\centering LLaMA Answers}} & \multirow{2}{*}{LLM-Based} & Mistral & 0.5160 & 0.6045 & 0.7704 & 0.4974 & 0.3731 & 0.1978 & 0.6930 & 0.1153 \\
 &  & LLaMA & 0.4940 & 0.6507 & 0.9818 & 0.4866 & 0.1674 & 0.2013 & 0.9408 & 0.1127 \\ \cline{2-11} 
 & {Text Classification}  & BERT & 0.664 & 0.664 & 0.651 & 0.650 & 0.532 & 0.377 & 0.034 & 0.808 \\ \cline{2-11} 
 &   & CC & 0.500 & 0.463 & 0.449 & 0.478 & 0.545 & 0.199 & 0.507 & 0.123 \\
 &  Specificity-based & DC & 0.484 & 0.464 & 0.465 & 0.463 & 0.562 & 0.190 & 0.462 & 0.120 \\
 &  QPP & IEF & 0.510 & 0.482 & 0.475 & 0.489 & 0.535 & 0.202 & 0.529 & 0.125 \\
 &  & PageRank & 0.482 & 0.461 & 0.461 & 0.461 & 0.534 & 0.151 & 0.371 & 0.094 \\ \cline{2-11} 
 & Supervised QPP  & BERTPE & 0.659 & 0.651 & 0.646 & 0.656 & 0.710 & 0.314 & 0.596 & 0.213 \\ \hline
 \hline
\end{tabular}}
\label{tab:baselines}
\end{table}

\noindent \textbf{Establishing Baselines.}  To benchmark this task, we identify three types of tasks from the literature that may be applicable to prompt sensitivity prediction: 

\noindent\textbf{(1)} We employ LLMs directly by asking them to self-assess their ability to predict whether they can accurately answer a given prompt or not~\cite{yan2024LLMevaluator}. 
This approach, which we refer to as the LLM self-evaluation baseline, relies on the model’s internal confidence in its responses. The prompts and instructions used for this baseline can be found in our GitHub repository for reproducibility.

\noindent\textbf{(2)} We further treat prompt sensitivity prediction as a text classification task. We train a text classifier on our dataset’s training set and evaluate it on the test set to predict whether the LLM’s response to a prompt will meet users' information need. We used the text classifiers implemented by Rajapakse et al. \cite{rajapakse2024simple}.

\noindent\textbf{(3) }
Prompt sensitivity prediction is also conceptually related to the task of query performance prediction \cite{poesina2024pqppjointbenchmarktexttoimage,salamat2023neural,arabzadeh2023noisy} whose goal is to estimate the quality of retrieved documents in response to a user query. For our task, we adapt QPP methods to predict whether a prompt will yield a correct response from an LLM or not.
QPP methods typically fall into pre-retrieval and post-retrieval categories. However, since generative settings do not produce traditional ``retrieved lists'', only pre-retrieval QPP methods are applicable in our context. Furthermore, collection-dependent QPP methods \cite{hauff2008survey} are also not applicable due to their dependence on a document corpus which is not available when using an LLM for generating a response. Thus, we adopted BERT-PE \cite{Khodabakhsh2024}, a SOTA pre-retrieval QPP model that similar to \cite{saleminezhad2024context,ebrahimi2024estimating}, uses contextualized embeddings to learn query performance. Additionally, we considered the neural embedding specificity-based QPP metrics to assess prompt sensitivity \cite{Arabzadeh2019,Arabzadeh2020,arabzadeh2020neural,arabzadeh2021query,arabzadeh2022unsupervised}, namely Closeness Centrality (CC), Degree Centrality (DC), PageRank , and Inverse Edge Frequency (IEF), all measured on the ego network of the query terms in the embedding space. We note that since the output of QPP methods is a scalar value, inspired by previous studies \cite{Faggioli2023,CollinsThompson2010,Meng2023}, we convert them to binary by classifying values above and below the mean of the data.


\begin{figure}[t]
\centering
\vspace{-2em}
\includegraphics[width=1\textwidth]{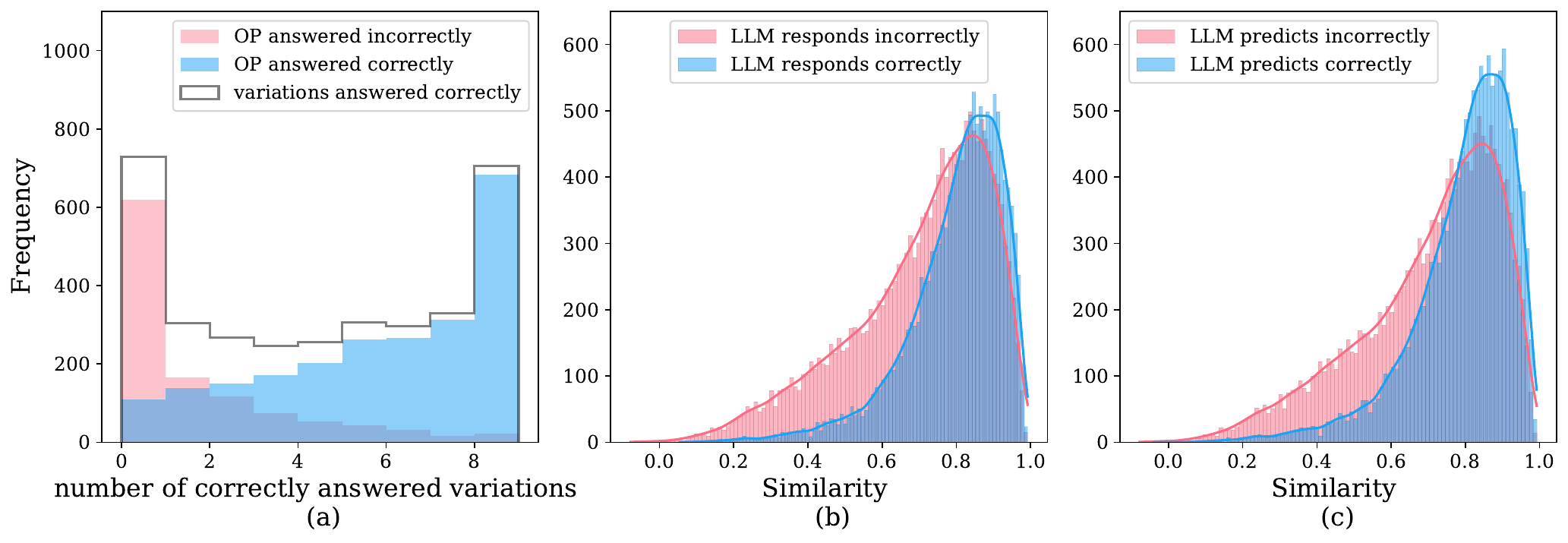} 
\caption{(a) A histogram of correctly answered variations when the original prompt yields a correct response (in blue) or an incorrect response (in red), along with the total count of correctly answered variations (black step). Figures (b) and (c) display histograms for correct (blue) and incorrect (red) responses and predictions, respectively. }
\label{fig:indepth1}
\end{figure}

\section{Experiments and Findings}

\noindent \textbf{Baseline Performance.} We analyze the performance of the of baselines on the \ds test set. Table \ref{tab:baselines} presents the results for predicting whether each prompt variation could be correctly answered using the LLaMA and Mistral. We report results in terms of accuracy, F1, recall, and precision. As shown in the table, the specificity-based QPP methods (i.e., CC, DC, IEF, and PageRank) perform the lowest among the baselines. Since QPP methods were not specifically designed for prompt sensitivity prediction, their performance is relatively weak on both datasets. We hypothesize that the specificity levels across the original prompts and their variations are too similar, making it challenging for specificity metrics to effectively distinguish between different levels of prompt specificity. On the other hand, BERT-PE demonstrates higher effectiveness in determining whether a prompt can be answered correctly. BERT-PE, which is supervised QPP method, shows competitive performance to text classification-based methods on \ds-TriviaQA and also outperforms other baselines significantly on \ds-HotpotQA. This suggests that \textit{supervised} QPP methods might be well-suited for the prompt sensitivity prediction task. We finally note the performance of LLM self-evaluation baseline. This baseline shows reasonable performance on TriviaQA but lacks consistency on HotpotQA. This is inverse to the performance of BERT-PE, indicating that these methods do not show stable performance across different prompt subsets.

\noindent \textbf{Impact of original prompt correctness on variation 
answerability\footnote{Due to space constraints, we report the full findings in our GitHub repo.}.} 
Here, we aim to investigate whether an LLM's ability to answer the original prompt influences its ability to answer other variations. Specifically, if the LLM can answer the original prompt, does this increase the likelihood of correctly answering the variations as well? To this end, Figure \ref{fig:indepth1}(a), marked with a black line, presents histograms of prompts showing the frequency of correctly answered variations out of a total of 10. In this figure, the x-axis represents the count of correctly answered variations for each prompt, grouping prompts by the number of successful variations. 
The distribution appears roughly balanced rather than long-tailed, indicating that \ds includes prompts with diverse answerability across reformulated variations, from those with only one answerable variation to those where all variations are answerable. In addition, in Figure \ref{fig:indepth1}(a), we further break down the results based on whether the original prompt was answered correctly or not. We observe that when the original prompt is answered correctly (shown in the blue histogram), there is a higher number of variations that also yield correct answers, reflected by an ascending pattern in the histogram. Conversely, when the original prompt is answered incorrectly, most of the prompts have only one out of the 10 variations answered correctly, displaying a descending pattern in the red histogram.

\begin{figure}[t]
\centering
\vspace{-2em}
\includegraphics[width=0.7\textwidth]{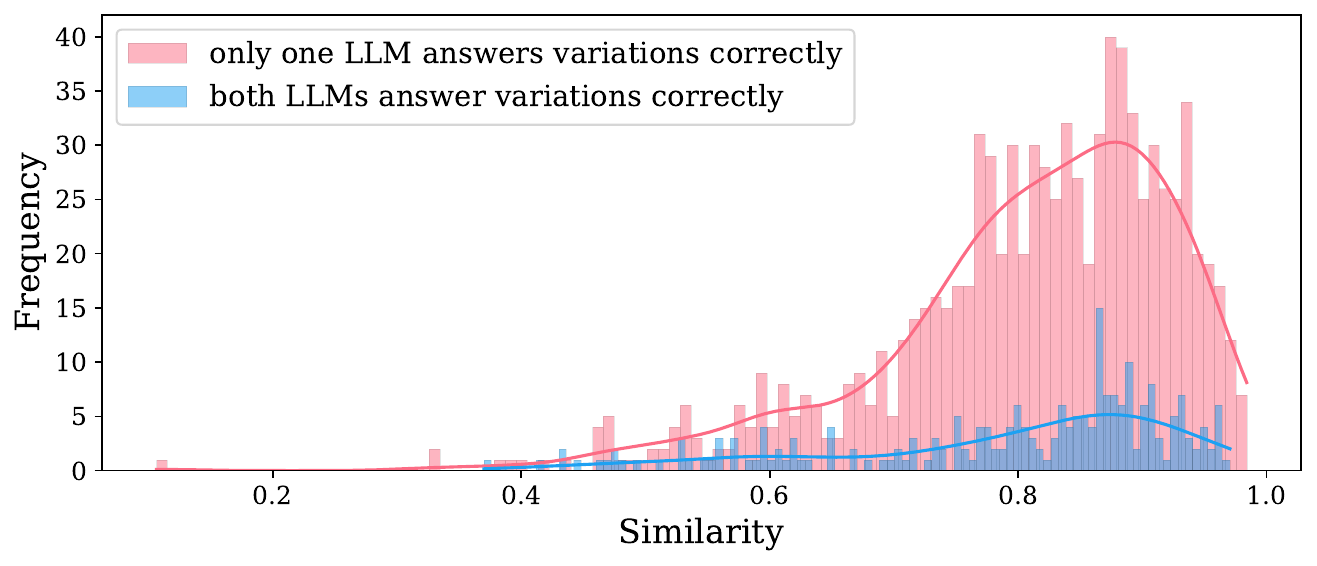} 
\caption{Distribution of answerability of variations of the questions in \ds test sets which both LLMs failed to answer correctly.  }
\label{fig:indepth2}
\end{figure}

\noindent \textbf{Impact of similarity to the original prompt. }Here, we aim to investigate the impact of variation similarity to the original prompt on the predictability of LLM responses. Specifically, we ask whether a variation that is more similar to the original prompt has a higher likelihood of generating a correct answer, while less similar variations may lead to lower predictability. To test this hypothesis, we conducted experiments, as shown in Figure \ref{fig:indepth1} (b) and (c).We present the distribution of correctly and incorrectly \textit{ answered prompts} and in Figure \ref{fig:indepth1}(b), and \textit{predicted responses} in Figure \ref{fig:indepth1}(c), based on similarity to the original prompt. Similarity is measured using the cosine similarity of the embedded representations of prompt-variation pairs, calculated with MiniLM, a model known for its strong performance in various NLP and IR tasks \cite{minilm}. We observe that when a variation closely resembles the original prompt, it is more likely to generate both correct responses and accurate predictions of answerability. This suggests that the model may have encountered this data points before, indicating a strong bias toward its training data and reduced generalizability to less familiar or novel prompt formulations.

\noindent \textbf{Impact of choice of LLM on variation answerability.}
We further explore whether prompt reformulation can enhance the effectiveness of an LLM. To investigate this, we first filter out questions from the \ds test set for which both LLMs, namely LLaMA and Mistral, failed to answer the original prompt correctly. Next, we examine the variations of these questions to see if an alternative prompt allows either LLM to provide a correct answer.
The results are shown in Figure \ref{fig:indepth2}. For each sample in this figure, both LLMs failed to answer the original prompt correctly. However, in the red cases, at least one of the two LLMs succeeded in answering a variation correctly, while in the blue cases, both LLMs provided correct answers to the variation. This highlights the potential of prompt reformulation as a strategy. We conclude that \ds can serve as a valuable resource for prompt reformulation, helping transform an unanswerable prompt into an answerable one through LLM-driven reformulation.

\section{Concluding Remarks}
This paper investigates the sensitivity of LLMs to prompt variations by introducing the Prompt Sensitivity Prediction task and the \ds dataset, based on TriviaQA and HotpotQA. We generate variations of different questions and examine the sensitivity of various LLMs to these variations, all of which share the same underlying information need. Our benchmarking results reveal that existing methods do not fully capture the complexities of prompt sensitivity. These findings underscore the need for further research into prompt variation sensitivity, particularly in developing methods to help users generate more reliable prompts.

\bibliographystyle{splncs04}
\bibliography{references}
\end{document}